\documentclass[10pt,twocolumn,letterpaper]{article}

\usepackage{cvpr}              
\definecolor{cvprblue}{rgb}{0.21,0.49,0.74}
\usepackage[pagebackref,breaklinks,colorlinks,allcolors=cvprblue]{hyperref}
\usepackage{listings}
\usepackage{algorithm}
\usepackage{multirow}
\usepackage{makecell}
\usepackage[table]{xcolor}
\usepackage{arydshln}
\usepackage{algpseudocode}
\usepackage{amsmath}
\usepackage{amssymb}
\usepackage{mathtools} 
\def\confName{CVPR}
\def\confYear{2026}

\title{GIFT: Global Irreplaceability Frame Targeting for Efficient Video Understanding}
\author{First Author\\
Institution1\\
Institution1 address\\
{\tt\small firstauthor@i1.org}
\and
Second Author\\
Institution2\\
First line of institution2 address\\
{\tt\small secondauthor@i2.org}
}

\author{
Junpeng Ma$^{1,2,4}$\thanks{These authors contributed equally to this work.}, 
Sashuai Zhou$^{3,4}$\footnotemark[1], 
Guanghao Li$^{1}$, 
Xin Gao$^{1}$,
Yue Cao$^{4}$,  
Hengyu Zeng$^{1}$, \\
Yuxiang Yan$^{1}$, 
Zhibin Wang$^{4,5}$\thanks{Project Leader.},
Jun Song$^{4,5}$, 
Bo Zheng$^{4,5}$, 
Shanghang Zhang$^{2}$\thanks{Corresponding Authors}, 
Jian Pu$^{1}$\footnotemark[3]\\
\small{$^1$Institute of Science and Technology for Brain-inspired Intelligence, Fudan University}\;\\
\small{$^2$State Key Laboratory of Multimedia Information Processing, School of Computer Science, Peking University}\;\\
\small{$^3$Zhejiang University}
\small{$^4$Alibaba Group Holding Limited}\;
\small{$^5$Future Living Lab of Alibaba}\;
}
\begin{document}
\maketitle 
\begin{abstract}
Video Large Language Models (VLMs) have achieved remarkable success in video understanding, but the significant computational cost from processing dense frames severely limits their practical application. Existing methods alleviate this by selecting keyframes, but their greedy decision-making, combined with a decoupled evaluation of relevance and diversity, often falls into local optima and results in erroneously selecting irrelevant noise frames. To address these challenges, we propose \textbf{GIFT}: \textbf{G}lobal \textbf{I}rreplaceability \textbf{F}rame \textbf{T}argeting, a novel training-free framework that selects frames by assessing their intrinsic irreplaceability. Specifically, we first introduce Directed Diversity to quantify a frame's uniqueness conditioned on relevance, which allows us to formulate a unified irreplaceability score. Subsequently, our Budget-Aware Refinement strategy employs a adaptive iterative process that first secures a core set of frames with the highest irreplaceability, and then shifts its priority to building crucial temporal context around these selections as the budget expands. Extensive experiments demonstrate that GIFT achieves a maximum average improvement of \textbf{12.5\%} across long-form video benchmarks on LLaVA-Video-7B compared to uniform sampling. Code is available at \url{https://github.com/ylnxxts/GIFT}.
\end{abstract}
\vspace{-3mm}    
\section{Introduction}
\label{sec:intro}
Recently, Video Large Language Models (VLMs)~\cite{chen2024expanding,li2024llava-ov,lin2023video-llava,song2024moviechat,li2024llamavid} have demonstrated impressive abilities in video understanding, achieving significant success across a wide range of tasks~\cite{chen2024vpl, chen2024multi, li2025constrained, li2025papl, gaodeep, Guo_2025_ICCV, wen2025Alpha-Service}. However, their powerful capabilities are often constrained by the massive number of visual tokens generated from dense video frames~\cite{zhang2025videollama3,chen2024longvila}. Processing these visual tokens results in substantial inference latency and memory consumption due to the quadratic complexity of the self-attention mechanism~\cite{vaswani2017transformer}, severely impeding their deployment in resource-constrained scenarios~\cite{han2026ficoco,liu2025vidcom2,liu2025globalcom2,cao2026fastdrivevla,liu2025mixkv,yan2024pointssc,yan2025ordermind,ma2025mmg,chen2025v2drop}. To mitigate this, most VLMs employ uniform sampling to reduce the number of input frames~\cite{li2023videochat,zhu2025internvl3,zhang2024llava-video,bai2025qwen2.5vl,shu2025videoxl}. However, this naive approach treats all frames equally, ignoring the fact that crucial information is often concentrated in a few key moments, inevitably leading to the inclusion of numerous redundant, query-irrelevant frames. This not only squanders the computational budget but also distracts the model's attention from critical information, thereby degrading its performance~\cite{liu2025bolt,sun2025f2c,cheng2025scaling}.
 \begin{figure}[t]
  \centering
   \includegraphics[width=1.0\linewidth]{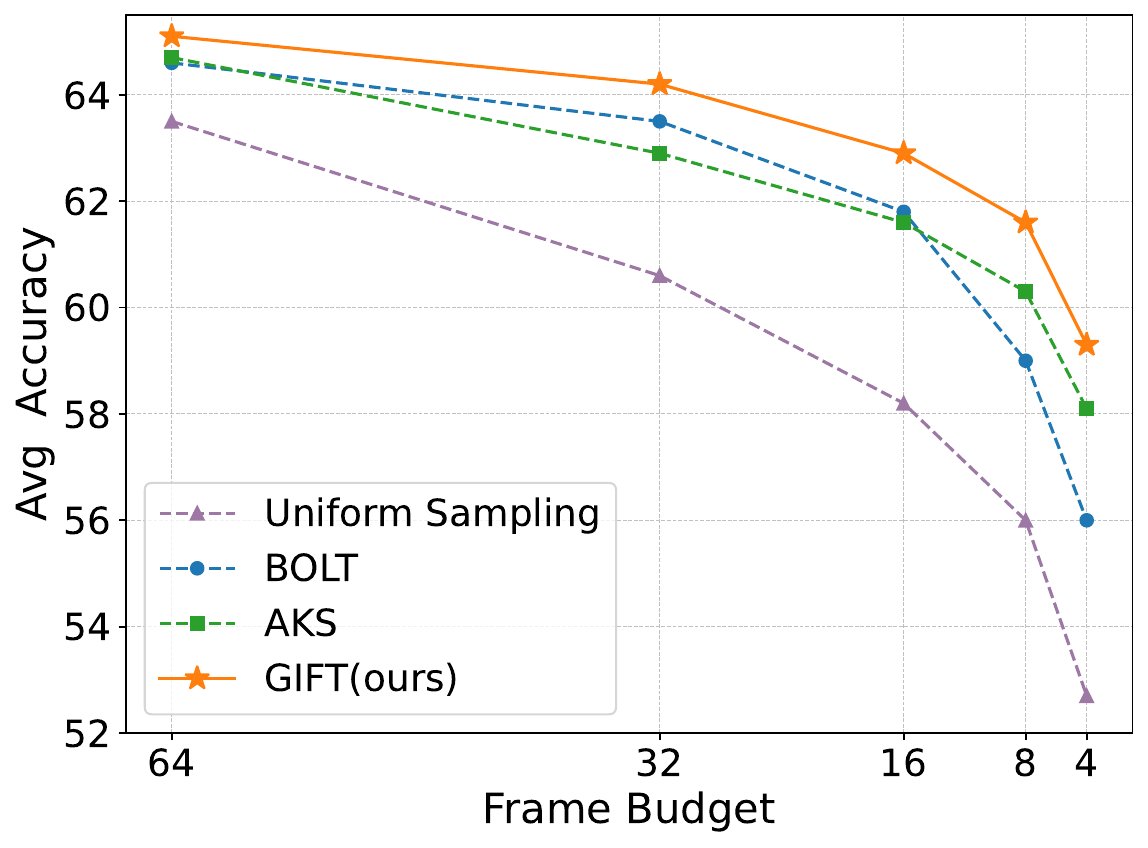}

   \caption{Average accuracy of LLaVA-Video-7B on Video-MME, LongVideoBench and MLVU for different methods: uniform sampling, BOLT~\cite{liu2025bolt}, AKS~\cite{tang2025aks}, and our GIFT.}
   \label{fig:intro}
\end{figure}

Consequently, developing effective keyframe selection algorithms to reduce redundancy while preserving the most task-critical information has become a central focus of current research~\cite{sun2025mdp3,tang2025aks,yu2024frame-voyager,wang2025videotree,wang2024videoagent}. We posit that this information is composed of three crucial aspects: (1) \emph{Query Relevance}, which equips the model with the direct visual evidence required to correctly answer a user's question; (2) \emph{Content Diversity}, which aims to maximize the coverage of useful information across the entire video within a limited budget; and (3) \emph{Temporal Coherence}, which captures the continuity of actions, as isolated frames are often insufficient for reasoning about dynamic events (e.g., \emph{identifying the player who scores a goal requires frames of the shooting motion, not just the ball crossing the line}). 

However, most existing methods fail to strike a meaningful balance for these components. Their struggles stem from two critical limitations in their design philosophy:

\noindent \textbf{(I) Myopia of Greedy-based Decisions:} Current approaches typically employ a greedy-based paradigm to solve the frame selection problem~\cite{zhu2025focus,guo2025logicframes,ye2025TreeSearch}. At each step, the algorithm makes a locally optimal and irrevocable choice based solely on the current state. This lack of a global perspective allows an early suboptimal decision to propagate and amplify its influence throughout the selection sequence, ultimately leading to a suboptimal solution.

\noindent \textbf{(II) Flawed Decoupled Criteria:} The fragility of the aforementioned process is worsened by a decoupled evaluation criterion, where query relevance and content diversity are treated as two independent objectives balanced by hand-tuned hyperparameters~\cite{sun2025f2c,tang2025aks}. The pursuit of diversity with this combination often sacrifices crucial temporal coherence while introducing irrelevant noise frames. More critically, when a suboptimal frame is incorrectly selected due to a minor diversity advantage, the truly optimal frame is subsequently and permanently excluded by the diversity mechanism for being too similar to the inferior choice.

To break free from this myopic and decoupled paradigm, we argue for a fundamental shift in perspective and propose a training-free keyframe selection framework ``\textbf{GIFT}: \textbf{G}lobal \textbf{I}rreplaceability \textbf{F}rame \textbf{T}argeting". Rather than asking ``\emph{What is the next best frame to add?}" to incrementally search for a locally optimal solution, our approach directly assesses the intrinsic irreplaceability of each frame by asking a powerful question: \textbf{\emph{Does a superior substitute exist?}} We define a superior substitute for a frame $F_i$ as any other frame $F_j$ that is both visually similar and more query-relevant, whose presence renders $F_i$'s contribution largely redundant. This principle allows us to directly quantify a frame's irreplaceability, forming a unified criterion for assessing frame importance from global perspective.

Specifically, our proposed GIFT consists of two core stages:
\textbf{(i) Quantifying Irreplaceability via Directed Diversity:} We formally define a frame's irreplaceability as the property of possessing high query-relevance while being visually remote from its potential substitutes set. To quantify this, we replace traditional diversity metrics with our proposed \emph{Directed Diversity} to measure dissimilarity exclusively against this set, which we define as all other frames with higher query relevance. A high irreplaceability score marks the frame as a unique, salient representation of task-critical information that should be prioritized for selection. \textbf{(ii) Budget-Aware Refinement:} While our irreplaceability criterion excels at identifying frames carrying the most essential information, its inherent suppression of adjacent frames can hinder tasks requiring temporal context. To resolve this, we introduce a Budget-Aware Refinement strategy that elegantly transitions the selection logic from a global assessment to an iterative refinement process. Initially, it prioritizes the frames with the highest irreplaceability scores. As the budget increases, the suppressive effect exerted by these selected keyframes is dynamically removed, allowing their once-suppressed but contextually vital neighbors to emerge as the next most salient choices. This iterative refinement naturally builds crucial temporal coherence and facilitates a more comprehensive understanding of key events.

We apply our GIFT to the most widely used VLMs, including LLaVA-Video~\cite{zhang2024llava-video}, LLaVA-OneVision~\cite{li2024llava-ov}, Qwen2.5-VL~\cite{bai2025qwen2.5vl}, and evaluate on multiple video question answering benchmarks~\cite{li2024mvbench,wu2024longvideobench,zhou2024mlvu,fu2024videomme}. \cref{fig:intro} presents the average results across three long-form video benchmarks on LLaVA-Video-7B. The results demonstrate that GIFT consistently outperforms all competing methods across all frame budgets, achieving a maximum average improvement of 12.5\% over uniform sampling.

The key contributions are summarized as follows:

\begin{itemize}
     \item We propose a training-free keyframe selection paradigm from a global optimization perspective. By introducing \emph{Directed Diversity} to quantify a frame's irreplaceability, our approach assesses each frame's value from a global, holistic standpoint, circumventing the error propagation pitfalls of greedy-based methods. 
     \item We propose a Budget-Aware Refinement strategy that dynamically adjusts the selection logic. This strategy transitions from prioritizing the most irreplaceable keyframes under low budgets to progressively incorporating their context as the budget increases, thereby providing the crucial temporal coherence for complex reasoning tasks.
     \item Our GIFT achieves state-of-the-art performance across various benchmarks. It can be seamlessly integrated with diverse VLMs, providing an effective solution for long-form video understanding.
\end{itemize}
\section{Related Work}
\label{sec:related_work}
\subsection{Video Large Language Models}
Recent studies have explored diverse paradigms to advance video understanding capabilities, which have demonstrated significant potential across a wide range of downstream tasks~\cite{zhou2026spatialrewardverifiablespatialreward,li2026ec,li2026artdeco,gaogood,wang2025stc,ding2026omnisift}. Addressing the challenge of processing lengthy videos, LongVA~\cite{zhang2024longva} introduces Long Context Transfer, enhancing long-video understanding by extending the LLM's context length.  For token efficiency, LongVU~\cite{shen2024longvu} proposes a spatiotemporal adaptive compression mechanism that effectively reduces tokens while preserving critical visual details, and LLaVA-OneVision~\cite{li2024llava-ov} utilizes bilinear interpolation for efficient token compression.  For enhancing fine-grained understanding, Qwen2-VL~\cite{wang2024qwen2vl} leverages M-RoPE to enhance the model's temporal awareness, while LLaVA-Video~\cite{zhang2024llava-video} introduces new-line tokens for spatio-temporal grounding. However, the performance of these powerful models is fundamentally constrained by the quality of the input frames~\cite{yao2025gens,he2025framethinker}, highlighting the critical need for an effective upstream frame selection strategy, which is the central focus of our work.
\subsection{Keyframe selection for VLMs}

While training-based keyframe selection~\cite{yu2024frame-voyager,buch2025ffs,yu2023sevila,hu2025m-llm-frameselect} approaches can be effective to select the most informative frames, their high computational cost and poorly scalability has motivated a shift towards training-free methods. Existing works focus on balancing query relevance with frame diversity. BOLT~\cite{liu2025bolt} employs inverse transform sampling to this end. AKS~\cite{tang2025aks} formulates the task as an optimization problem over relevance and video coverage. MaxInfo~\cite{li2025maxinfo} first reduces the dimensionality of frame embeddings via SVD, and then utilizes the maximum volume principle within this subspace to select a maximally representative and diverse subset of keyframes. MDP3~\cite{sun2025mdp3} emphasizes the joint optimization of query relevance, list-wise diversity, and sequentiality through a combination of Determinantal Point Processes and Markov Decision Processes. Most recently, Q-Frame~\cite{zhang2025qframe} introduces dynamic resolution by ranking frames into multiple resolution levels, thereby reducing visual tokens. However, their approach of treating relevance and diversity as separate objectives often relies on greedy-based logic, making it susceptible to local optima. In contrast, our method provides a unified and global selection criterion by redefining diversity to be conditional on relevance,  while employing a dynamic selection process that adapts to the available frame budget.
\section{Method}

\label{sec:method}
 \begin{figure*}[t]
  \centering
   \includegraphics[width=\linewidth]{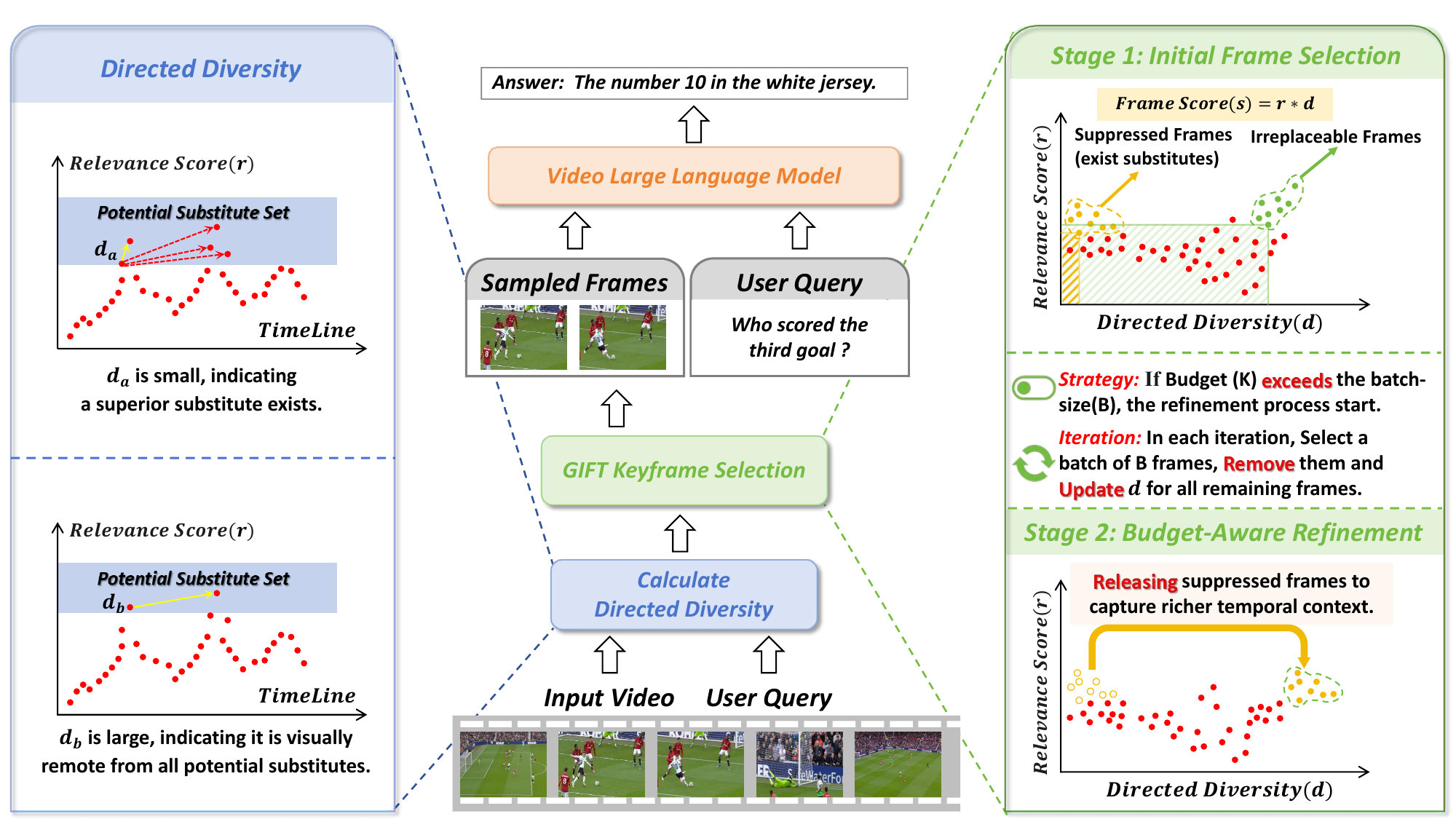}
   \caption{ \textbf{The overall framework of our GIFT.} Given an input video and user query, GIFT first calculates each frame's query-relevance($r$) and directed diversity($d$) to quantify its irreplaceability(detailed in the left panel). The core of GIFT is a two-stage selection process (detailed in the right panel).  Stage 1 performs the initial selection by identifying frames with high scores ($s=r \times d$). If the sampling budget ($K$) exceeds batch-size ($B$), the Budget-Aware Refinement in Stage 2 is triggered. This stage iteratively Selects a batch of $B$ frames, \emph{Removes} them, and \emph{Updates} the $d$ for all remaining frames until the budget is met. This iterative update process continuously \emph{Releases} previously suppressed frames, progressively building a rich temporal context.}
   \label{fig:method}
\end{figure*}




In this section, we first revisit the conventional formulation of the keyframe selection task to highlight its inherent limitations. We then propose a new problem formulation centered around our core principle of global irreplaceability, setting the stage for our proposed framework, GIFT. 
\subsection{Problem Formulation}
Conventionally, keyframe selection is formulated as a constrained subset selection problem. Given a video $\mathcal{F}_{\text{v}} = \{F_1, \dots, F_N\}$, the goal is to find a subset $\mathcal{F}^*_{\text{sub}}$ of size $K$ that maximizes query relevance and subset diversity: 
\begin{equation} \label{eq:1}
    \mathcal{F}^*_{sub} = \mathop{\mathrm{arg\,max}}_{\mathcal{F} \subset \mathcal{F}_{\text{v}}, |\mathcal{F}|=K}  \underbrace{\sum_{f_i \in \mathcal{F}} \text{sim}(f_i, q)}_{\text{Total Relevance}} - \lambda \underbrace{\sum_{\substack{f_i, f_j \in \mathcal{F} \\ i \neq j}} \text{sim}(f_i, f_j)}_{\text{Total Redundancy}},
\end{equation}
where $f_i$ is the visual embedding of frame $F_i$ and $q$ is the text embedding of query, both extracted from a VLM. As discussed in ~\cref{sec:intro}, this formulation has several limitations. Its NP-hard nature often necessitates myopic, greedy-based solutions that suffer from error propagation. More critically, it treats relevance and diversity as decoupled, competing objectives, leading to the destructive trade-offs.

To overcome these limitations, we re-formulate the problem entirely. We argue that the optimal objective is not to balance two separate metrics, but to select a subset of frames that maximizes a single, unified property: total irreplaceability. Our proposed objective is therefore:

\begin{equation}
    \label{eq:2}
    \mathcal{F}^*_{\text{sub}} = \mathop{\mathrm{arg\,max}}_{\mathcal{F} \subset \mathcal{F}_{\text{v}}, |\mathcal{F}|=K} \sum_{F_i \in \mathcal{F}} \text{Irreplaceability}(F_i, Q, \mathcal{F}_{\text{v}}) .
\end{equation}

Here, $\text{Irreplaceability}(F_i, Q, \mathcal{F}_{\text{v}})$ is a global score that quantifies the unique, task-critical information contributed by frame $F_i$ in the context of the query $Q$ and the entire video $\mathcal{F}_{\text{v}}$. The core of our method thus becomes the definition and computation of this irreplaceability score. Our GIFT addresses this challenge through the two synergistic stages: \emph{Quantifying Irreplaceability via Directed Diversity} and \emph{Budget-Aware Refinement}. The following subsections will detail the specifics of each stage.

\subsection{Quantifying Irreplaceability via Directed Diversity}

We define a frame's irreplaceability as a holistic measure that operationalizes our principle of identifying frames for which no \emph{``superior substitute exists"}. A superior substitute is any other frame that is both visually similar and more query-relevant, whose existence renders the original frame's informational contribution largely redundant.

To translate this principle into a computable score, we factorize the irreplaceability score for each frame $F_i$ into two components: its fundamental relevance to the query and its uniqueness against potential alternatives.

\noindent \textbf{(i) Query Relevance.} 
This component measures the direct semantic alignment between a frame $F_i$ and the user's query $Q$, serving as the foundational measure of a frame's merit. We define it as the cosine similarity between their respective embeddings $f,q$:
\begin{equation}
\label{eq:3}
\begin{aligned}
    r'_i &= \text{sim}(f_i, q) = \frac{f_i \cdot q}{\|f_i\| \|q\|} , \\
    r_i &= \frac{r'_i - \min(\mathcal{R}')}{\max(\mathcal{R}') - \min(\mathcal{R}')} \:.
\end{aligned}
\end{equation}
\textbf{(ii) Directed Diversity.} 
This is the key innovation of our framework, fundamentally redefining the concept of diversity. Unlike traditional metrics that measure a frame's dissimilarity to other frames, our directed diversity is explicitly unidirectional. It quantifies a frame's uniqueness \emph{conditioned on relevance} by measuring its minimum distance to any frame within its set of ``potential substitutes''. Conceptually, we define this set as all other frames that are more query-relevant. The formal definition is given below:

\begin{equation}
\label{eq:4}
\begin{aligned}
    & d_i =
    \begin{cases}
        \displaystyle \min_{j \in \mathcal{C}_i}  \|f_i - f_j\|_{2}^{2}, & \text{if } \mathcal{C}_i \neq \emptyset \\
        \displaystyle \max_{F_j, F_k \in \mathcal{F}_{\text{v}}} \|f_j - f_k\|_{2}^{2}, & \text{otherwise}
    \end{cases}
    \\ 
    & \text{where} \quad 
        \mathcal{C}_i = \{ j \in \{1,\dots,N\}  \mid r_j > r_i \} .
\end{aligned}
\end{equation}
Here, $\mathcal{C}_i$ is the set of potential substitutes and $d_i$ is the directed diversity score for $F_i$. A special case arises when this set is empty, which indicates that $F_i$ is the most query-relevant frame in the entire video. To reflect its unique status of having no potential substitutes, we assign it the maximum possible distance in the embedding space as its directed diversity score.
This conditional formulation has a powerful discerning effect:

\textbf{A low $d_i$} acts as a strong penalty for redundancy. It signifies that a superior substitute $F_j$ exists in close visual feature space, meaning the unique contribution of $F_i$, encompassing both its visual content and query-related information, is largely subsumed by this substitute.

\textbf{A high $d_i$} validates a frame's uniqueness among its potential alternatives. This occurs under two scenarios:
\begin{itemize}
    \item The frame is the most query-relevant in the entire video ($\mathcal{C}_i = \emptyset$), leaving no alternatives to challenge its primacy and resulting in a maximal diversity score. 
    \item While more query-relevant frames exist ($\mathcal{C}_i \neq \emptyset$), all of them are visually remote from $F_i$. This high distance indicates that the frame represents distinct visual information not captured by any of its potential substitutes, making it a valuable candidate for selection.
\end{itemize}

\noindent \textbf{(iii) Final Irreplaceability Score.} Finally, we compute the irreplaceability score $s$ by multiplying the two components: 
\begin{equation}
\label{eq:5}
    s_i = r_i \times d_i \:.
\end{equation}

This formulation ensures that only frames which are both highly relevant (high $r_i$) and unique against potential alternatives (high $d_i$) receive the highest priority. Since each frame is assigned a static irreplaceability score based on this global assessment, the combinatorial optimization problem of maximizing the total subset score (as formulated in \cref{eq:2}) simplifies to a straightforward selection of the $K$ frames with the highest scores.

\subsection{Budget-Aware Refinement}

Our global irreplaceability score is predicated on a principle:\emph{ a small subset of the most irreplaceable frames is sufficient to represent the video's core, task-critical information}. While this principle is highly effective for identifying the vital  moments, such a static, one-off scoring mechanism faces inherent limitations, particularly under two conditions:

\begin{enumerate}
    \item \textbf{Task Mismatch in Temporal Reasoning:} 
   The framework's core logic of maximizing informational uniqueness directly conflicts with the need for temporal coherence in certain tasks. While our method excels at identifying the primary task-critical frames, its intrinsic suppression of visually similar frames inherently penalizes the selection of temporally adjacent, causally linked frames. This is particularly problematic for fine-grained temporal reasoning, such as analyzing the continuous motion of a goal-scoring sequence.
       
    \item \textbf{Quality Degradation with Increasing Budgets}: This issue is worsened as the frame budget increases. The static nature of the irreplaceability score dictates that frames temporally adjacent to the primary, high-scoring selections are often heavily suppressed due to their inherent visual similarity. Consequently, once these primary frames have been chosen, the algorithm may be compelled to select isolated, low-relevance noise frames that have a spuriously high directed diversity score over more query-relevant, yet suppressed, secondary frames.
\end{enumerate}

\noindent These challenges reveal that a single, static assessment of frame importance is insufficient. A truly robust framework should not only identify the most critical frames but also dynamically refine its selection strategy based on the available budget. To this end, we introduce a Budget-Aware Refinement strategy. The core principle of this stage is to transition the selection focus based on budget constraints: at low budgets, the primary priority should be to maximize coverage of task-critical information. As the budget increases, the priority should then naturally pivot towards enriching the temporal context around these established keyframes to capture the complete narrative flow of key events.

Our GIFT elegantly realizes this transition through an iterative \emph{select, remove, and re-evaluate} process, detailed in \cref{alg:1}. In each iteration, a small batch of $B$ frames with the highest current irreplaceability scores is chosen. Crucially, these frames are then removed from the candidate pool, and the irreplaceability scores for all remaining frames are re-evaluated before the next iteration begins. 
\begin{algorithm}[t]
\caption{GIFT keyframe selection}
\label{alg:1}
\begin{algorithmic}[1]
\State \textbf{Input:} Set of frame embeddings $F = \{f_i\}_{i=1}^N$; Query embedding $q$; Frame Budget $K$; Batch size $B$.
\State \textbf{Output:} A set of selected frame indices $\mathcal{I}^*_{sub}$ with $|\mathcal{I}^*_{sub}| = K$.

\Statex \textit{\# Initialization }
\State $\mathcal{I}^*_{sub} \gets \emptyset$
\State $\mathcal{I}_{\text{cand}} \gets \{1, 2, \dots, N\}$ \quad \textit{\# Set of candidate indices}

\Statex \textit{\# 1. Pre-compute Query-Relevance  }
\For{each index $i \in \mathcal{I}_{\text{cand}}$}
    \State $r'_i \gets \text{cosine\_similarity}(f_i, q)$
\EndFor
\State \textbf{Normalize all $r'_i$ to get $r$ using \cref{eq:3}}
\State $d_{max} \gets \max\limits_{i,j \in \mathcal{I}_{\text{cand}}} \text{dist}(f_i, f_j)$
    
\While{$|\mathcal{I}^*_{sub}| < K$}

    \For{each candidate index $i \in \mathcal{I}_{\text{cand}}$}
        \Statex \hspace{\algorithmicindent} \textit{\# 2. Compute Directed Diversity}
        \State     $\mathcal{C}_i \gets \{ j \in \mathcal{I}_{\text{cand}} \mid r_j > r_i \}$
        
        \State $d_i \gets  \begin{cases} \min\limits_{j \in \mathcal{C}_i} \text{dist}(f_i, f_j) & \text{if } \mathcal{C}_i \neq \emptyset \\ d_{max} & \text{otherwise}  \end{cases}$       
        \Statex \hspace{\algorithmicindent} \textit{\# 3. Calculate Irreplaceability Score}
        \State $s_i \gets r_i \times d_i$
    \EndFor
    
    \Statex \hspace{\algorithmicindent} \textit{\# 4. Select a Batch and Update Candidate Sets}
    \State $b \gets \min(B, K - |\mathcal{I}^*_{sub}|)$
    \State $\mathcal{B} \leftarrow \underset{i \in \mathcal{I}_{\text{cand}}}{\text{argmax}_b} (s_i)$
    
    \State $\mathcal{I}^*_{sub} \gets \mathcal{I}^*_{sub} \cup \mathcal{B}$
    \State $\mathcal{I}_{\text{cand}} \gets \mathcal{I}_{\text{cand}} \setminus \mathcal{B}$
\EndWhile

\State \Return $\mathcal{I}^*_{sub}$
\end{algorithmic}
\end{algorithm}
This iterative re-evaluation inherently achieves the desired adaptive behavior. The removal of selected keyframes in each stage eliminates the powerful suppression they exert on their neighbors, which allows the framework to enrich the selection with crucial temporal coherence as the budget increases. For instance, consider a goal-scoring sequence where the frame of the ball crossing the goal line is selected first due to its salient query-relevance. Without this dynamic adjustment, other frames crucial to the action would be suppressed by this initial selection, making it impossible for the model to capture the complete motion. 
\begin{table*}[t]
\caption{\textbf{Comparison of state-of-the-art methods across video understanding benchmarks on LLaVA-Video-7B.}  Our GIFT consistently outperforms competing approaches across all frame budgets, demonstrating superior performance especially under tight budget constraints. \textbf{Best} results are highlighted in bold, \underline{second best} underlined.}
\label{tab:1}
\centering
\renewcommand{\arraystretch}{1.05}
\setlength{\tabcolsep}{1mm}
{
\begin{tabular}{c|ccccccc|cc}
\hline
\multirow{2}{*}{\textbf{Method}} & \multirow{2}{*}{\textbf{MVBench}} & \multirow{2}{*}{\makecell{\textbf{LongVideo}\\\textbf{Bench}}} & \multirow{2}{*}{\textbf{MLVU}} & \multicolumn{4}{c}{\textbf{VideoMME(w.o. sub.)}} & \multicolumn{2}{|c}{\multirow{2}{*}{Average}} \\
\cline{5-8}
& & & & Overall & Short & Medium & Long & & \\
Duration & 16s & 1$\sim$60min & 3$\sim$120min & 1$\sim$60min & 1$\sim$3min & 3$\sim$30min & 30$\sim$60min & Score & \% \\ \hline
LLaVA-Video(64Frames)                      & 59.2& 58.9& 67.4& 64.3& 77.3& 62.3& 53.2&  62.5& 100.0\\  \hline
\multicolumn{10}{c}{\textit{\textbf{Frame Budget: 32Frames}}} \\ \hline

Uniform Sampling& \underline{59.2}& 58.0& 61.2& 62.6& \underline{76.3}& 59.2& 52.3& 60.3
& 96.5\\
BOLT{\small\texttt{(CVPR2025)}} & \textbf{59.7}& \underline{59.8}& \underline{66.5}& \underline{64.1}& 75.7& 62.7& \textbf{54.0}& \underline{62.5}& \underline{100.0}\\
AKS{\small\texttt{(CVPR2025)}}& 58.5& 59.6& 65.4& 63.6& 75.1& \underline{63.9}& 51.7& 61.8
& 98.9\\
\rowcolor{gray!15}
GIFT{\small\texttt{(ours)}}& 59.1& \textbf{60.2}& \textbf{67.4}& \textbf{65.0}& \textbf{77.3}& \textbf{64.1}& \underline{53.6}& \textbf{62.9}& \textbf{100.6}\\ \hline
\multicolumn{10}{c}{\textit{\textbf{Frame Budget: 16Frames}}} \\ \hline

Uniform Sampling
& \textbf{59.1}& 57.1& 57.6& 59.9& 70.7& 58.7& 50.3& 58.4
& 93.4\\
BOLT{\small\texttt{(CVPR2025)}}& \underline{59.0}& \underline{58.6}& \underline{64.8}& 62.0& \underline{75.4}& 60.3& 50.3& \underline{61.1}& \underline{97.8}\\
AKS{\small\texttt{(CVPR2025)}}& 57.3& \textbf{58.9}& 63.9& \underline{62.1}& 72.6& \underline{61.6}& \underline{52.1}& 60.6
& 97.0\\
\rowcolor{gray!15}
GIFT{\small\texttt{(ours)}}& \textbf{59.1}& \textbf{58.9}& \textbf{66.1}& \textbf{63.8}& \textbf{76.0}& \textbf{61.9}& \textbf{53.6}& \textbf{62.0}& \textbf{99.2}\\ \hline
\multicolumn{10}{c}{\textit{\textbf{Frame Budget: 8Frames}}} \\ \hline

Uniform Sampling
& \underline{58.1}& 55.7& 56.4& 56.0& 67.7& 53.6& 46.7& 56.6
& 90.6\\
BOLT{\small\texttt{(CVPR2025)}}
& 57.7& 56.3& 61.2& \underline{59.5}& \underline{73.0}& 56.2& 49.3& 58.7
& 93.9\\
AKS{\small\texttt{(CVPR2025)}}
& 56.1& \underline{57.4}& \underline{64.1}& 59.4& 68.8& \underline{58.6}& \underline{51.0}& \underline{59.3}& \underline{94.9}\\
\rowcolor{gray!15}
GIFT{\small\texttt{(ours)}}& \textbf{58.2}& \textbf{57.5}& \textbf{65.8}& \textbf{61.5}& \textbf{74.0}& \textbf{59.1}& \textbf{51.3}& \textbf{60.8}& \textbf{97.3}\\ \hline
\multicolumn{10}{c}{\textit{\textbf{Frame Budget: 4Frames}}} \\ \hline

Uniform Sampling
& \underline{56.1}& 51.8& 52.9& 53.3& 62.6& 50.2& 47.0& 53.5
& 85.6\\
BOLT{\small\texttt{(CVPR2025)}}
& 55.9& 53.9& 57.4& 56.6& \underline{67.2}& 55.0& 47.6& 56.0
& 89.6\\
AKS{\small\texttt{(CVPR2025)}}
& 54.6& \underline{55.3}& \underline{61.7}& \underline{57.4}& 66.2& \textbf{57.0}& \textbf{49.1}& \underline{57.3}& \underline{91.7}\\
\rowcolor{gray!15}
GIFT{\small\texttt{(ours)}}& \textbf{57.0}& \textbf{56.2}& \textbf{63.1}& \textbf{58.6}& \textbf{69.8}& \underline{56.9}& \underline{49.0}& \textbf{58.7}& \textbf{93.9}\\ 
\hline
\end{tabular}
}
\vspace{-1mm}
\end{table*}

In contrast, our adaptive approach directly counters this by removing the already-selected keyframe from the candidate pool when the budget allows, which in turn releases its suppressive effect. This allows the once-suppressed yet contextually vital neighbors to be captured in subsequent iterations. Consequently, GIFT naturally transitions from capturing the single most critical moment to reconstructing the entire event narrative as the budget expands. Furthermore, this process is inherently robust to noise, as noisy frames consistently receive low irreplaceability scores due to their poor alignment with the query.

\section{Experiments}
\label{sec:experiments}

\subsection{Experimental Settings}
\noindent \textbf{Benchmarks and Baselines:}
We employ lmms-eval~\cite{zhang2024lmms-eval} to evaluate our method on four widely-used video understanding benchmarks: MVBench~\cite{li2024mvbench}, LongVideoBench~\cite{wu2024longvideobench}, MLVU~\cite{zhou2024mlvu}, and VideoMME~\cite{fu2024videomme}. These benchmarks encompass a wide range of video durations and a diversity of complex scenarios, enabling a comprehensive assessment of our method's effectiveness and generalizability. For comparison, we compare our method against uniform sapmling and two representative open-source methods, including BOLT~\cite{liu2025bolt} and AKS~\cite{tang2025aks}. Uniform Sampling selects frames at equal temporal intervals across the video, serving as a common baseline. BOLT utilizes Inverse Transform Sampling to probabilistically select frames from a cumulative distribution function defined by query relevance scores, thus prioritizing relevant content while preserving temporal diversity. AKS employs a recursive partitioning algorithm to adaptively select frames, optimizing a trade-off between query relevance and temporal coverage.
\definecolor{green}{HTML}{34A853}
\newcommand{\gainfoot}[1]{\scriptsize\color{green}\textbf{#1$\uparrow$}}
\newcommand{\emptyfoot}[1]{\scriptsize\textbf{\phantom{#1$\uparrow$}}}

\begin{table*}[t]
\caption{\textbf{Performance comparison of state-of-the-art VLMs with and without our GIFT enhancement.} GIFT delivers consistent and significant performance gains across all base models, benchmarks and frame budgets (8 and 32), demonstrating its strong generalizability as a model-agnostic, plug-and-play solution.}
\label{tab:2}
\centering
\tabcolsep=0.16cm
\renewcommand{\arraystretch}{1.05}
\begin{tabular}{lcccccc}
    \toprule
		\multirow{2}{*}{\textbf{Model}} & \multirow{2}{*}{\textbf{LongVideoBench}}& \multirow{2}{*}{\textbf{MLVU\emptyfoot{+0.0}}}& \multicolumn{4}{c}{\textbf{VideoMME(w.o. sub.)}} \\
        \cline{4-7} & & & Overall\emptyfoot{+0.0} & Short\emptyfoot{+0.0} & Medium\emptyfoot{+0.0} & Long\emptyfoot{+0.0}  \\
        \midrule
        VILA-V1.5$^{\text{8 frame}}$ & 47.1\emptyfoot{+0.0} & 49.8\emptyfoot{+0.0} & 48.4\emptyfoot{+0.0} & 59.4\emptyfoot{+0.0} & 44.7\emptyfoot{+0.0} & 41.1\emptyfoot{+0.0}  \\
        \rowcolor{gray!15}
        \textbf{VILA-V1.5$^{\text{8 frame}}$+GIFT} & \textbf{50.9\gainfoot{+3.8}} & \textbf{56.7\gainfoot{+6.9}} & \textbf{52.8\gainfoot{+4.4}} & \textbf{65.7\gainfoot{+6.3}} & \textbf{49.8\gainfoot{+5.1}} & \textbf{42.9\gainfoot{+1.8}} \\
	LLaVA-OneVision$^{\text{8 frame}}$ & 54.3\emptyfoot{+0.0} & 58.5\emptyfoot{+0.0} & 53.9\emptyfoot{+0.0} & 64.9\emptyfoot{+0.0} & 52.0\emptyfoot{+0.0} & 44.7\emptyfoot{+0.0} \\
        \rowcolor{gray!15}
        \textbf{LLaVA-OneVision$^{\text{8 frame}}$+GIFT} & \textbf{59.6\gainfoot{+5.3}} & \textbf{67.3\gainfoot{+8.8}} & \textbf{58.8\gainfoot{+4.9}} & \textbf{69.9\gainfoot{+5.0}} & \textbf{57.9\gainfoot{+5.9}} & \textbf{48.7\gainfoot{+4.0}} \\
        Qwen2.5-VL$^{\text{8 frame}}$ & 52.7\emptyfoot{+0.0} & 53.8\emptyfoot{+0.0} & 53.6\emptyfoot{+0.0} & 62.7\emptyfoot{+0.0} & 50.7\emptyfoot{+0.0} & 47.6\emptyfoot{+0.0} \\
        \rowcolor{gray!15}
        \textbf{Qwen2.5-VL$^{\text{8 frame}}$+GIFT} & \textbf{58.3\gainfoot{+5.6}} & \textbf{62.8\gainfoot{+9.0}} & \textbf{58.1\gainfoot{+4.5}} & \textbf{68.4\gainfoot{+5.7}} &   \textbf{56.9\gainfoot{+6.2}} & \textbf{49.0\gainfoot{+1.4}} \\
        VideoLLaMA3$^{\text{8 frame}}$& 54.8\emptyfoot{+0.0} & 59.1\emptyfoot{+0.0} & 59.1\emptyfoot{+0.0} & 70.0\emptyfoot{+0.0} & 56.4\emptyfoot{+0.0} & 50.9\emptyfoot{+0.0} \\
        \rowcolor{gray!15}
        \textbf{VideoLLaMA3$^{\text{8 frame}}$+GIFT}& \textbf{59.2\gainfoot{+4.4}} & \textbf{ 70.7\gainfoot{+11.6}} & \textbf{63.6\gainfoot{+4.5}} & \textbf{75.9\gainfoot{+5.9}} & \textbf{62.0\gainfoot{+5.6}} & \textbf{53.0\gainfoot{+2.1}} \\
        \hdashline
        LLaVA-OneVision$^{\text{32 frame}}$ & 56.6\emptyfoot{+0.0} & 63.1\emptyfoot{+0.0} & 58.5\emptyfoot{+0.0} & 70.1\emptyfoot{+0.0} & 56.6\emptyfoot{+0.0} & 48.8\emptyfoot{+0.0} \\  
        \rowcolor{gray!15}
        \textbf{LLaVA-OneVision$^{\text{32 frame}}$+GIFT} & \textbf{59.6\gainfoot{+3.0}}& \textbf{69.4\gainfoot{+6.3}} &\textbf{61.2\gainfoot{+2.7}} & \textbf{73.3\gainfoot{+3.2}} & \textbf{59.6\gainfoot{+3.0}} & \textbf{50.8\gainfoot{+2.0}} \\
        Qwen2.5-VL$^{\text{32 frame}}$ & 59.5\emptyfoot{+0.0} & 59.6\emptyfoot{+0.0} & 61.0\emptyfoot{+0.0} & 71.7\emptyfoot{+0.0} & 60.1\emptyfoot{+0.0} & 51.1\emptyfoot{+0.0} \\
        \rowcolor{gray!15}
        \textbf{Qwen2.5-VL$^{\text{32 frame}}$+GIFT} & \textbf{61.3\gainfoot{+1.8}} & \textbf{68.2\gainfoot{+8.6}} &\textbf{64.4\gainfoot{+3.4}} & \textbf{76.8\gainfoot{+5.1}} & \textbf{63.2\gainfoot{+3.1}} & \textbf{53.1\gainfoot{+2.0}} \\

        VideoLLaMA3$^{\text{32 frame}}$& 58.3\emptyfoot{+0.0} & 62.5\emptyfoot{+0.0} & 64.0\emptyfoot{+0.0} & 77.1\emptyfoot{+0.0} & 61.4\emptyfoot{+0.0} & 53.4\emptyfoot{+0.0} \\
                \rowcolor{gray!15}
        \textbf{VideoLLaMA3$^{\text{32 frame}}$+GIFT} & \textbf{58.9\gainfoot{+0.6}} & \textbf{68.2\gainfoot{+5.7}} & \textbf{66.4\gainfoot{+2.4}} &\textbf{79.8\gainfoot{+2.7}} & \textbf{64.2\gainfoot{+2.8}} & \textbf{55.1\gainfoot{+1.7}} \\
        \bottomrule
	\end{tabular}
\vspace{-1mm}
\end{table*}

\noindent \textbf{Implementation Details:} 
GIFT is implemented into several VLMs, including LLaVA-Video-7B~\cite{zhang2024llava-video}, VILA-V1.5-8B~\cite{lin2024vila}, LLaVA-OneVision-7B~\cite{li2024llava-ov}, Qwen2.5-VL-7B~\cite{bai2025qwen2.5vl} and VideoLLaMA3-7B~\cite{zhang2025videollama3}, to comprehensively assess its generalizability across diverse model architectures. For computational efficiency, we uniformly sample 128 candidate frames to apply our method. To ensure a fair comparison, we utilize the pretrained SigLIP~\cite{zhai2023siglip} model to extract visual and text embeddings for computing the similarity between video frames and queries across all methods. Furthermore, we set $B=9$ for all experiments. For BOLT, the hyperparameter $\alpha$ is set to 2.5. For AKS, the max depth $L$ is set to 3 and the threshold $s_{thr}$ is set to 0.2. All experiments are conducted on NVIDIA H100-SXM5-80GB GPUs.
\subsection{Main Results}
\cref{tab:1} presents a comparative analysis of our GIFT against several keyframe selection methods across various benchmarks. \cref{tab:2} details the performance of GIFT when applied to diverse VLMs. The experimental results reveal three key advantages of GIFT:

\noindent \textbf{(i) State-of-the-art Performance:} As detailed in \cref{tab:1}, we established four distinct frame budgets \{32, 16, 8, 4\}  to thoroughly assess the performance of GIFT and evaluate its robustness under varying budget constraints. The experimental results demonstrate that GIFT consistently achieves optimal performance across all configurations, significantly outperforming all other baseline methods. For instance, at a budget of 32 frames, GIFT achieves an average score of 62.9, surpassing the Uniform sampling by 2.6(+4.3\%). This consistent superiority validates our core framework's ability to identify a more globally optimal and task-relevant frame subset compared to myopic, greedy-based approaches.

\noindent \textbf{(ii) Superior Performance under Severe Budget Constraints:} GIFT's advantage is particularly pronounced under tight budgets, where our method's performance degradation is substantially less pronounced than that of other methods. With only 4 frames, GIFT retains 93.9\% of the LLaVA-Video(64frames) performance, a remarkable 8.3\% higher than Uniform Sampling, and outperforms the best baseline by 2.2\%. We attribute this to our global selection strategy which effectively identifies \emph{irreplaceability frame}, making it significantly more robust to the redundant and noisy frames(e.g., static scenes, blurry shots) than myopic, greedy-based approaches. Furthermore, our method's strength extends to tasks where temporal coherence is critical. In short videos, GIFT demonstrates a superior ability to preserve the crucial context around key events, as it does not over-penalize adjacent frames in pursuit of visual diversity. For example, on the VideoMME-Short benchmark, GIFT achieves a score of 74.0 with 8 frames and 69.8 with 4 frames, surpassing the second best methods by 1.4\% and 3.9\%, respectively. 

\noindent \textbf{(iii) Cross-Model Robustness:} To validate the cross-model generalizability and plug-and-play capability of our proposed GIFT, we integrated it into four diverse VLMs: VILA-V1.5, LLaVA-OneVision, Qwen2.5-VL, and VideoLLaMA3. As presented in \cref{tab:2}, GIFT consistently yields significant performance gains, regardless of the underlying model architecture or the frame budget. For instance, when applied to LLaVA-OneVision with a sparse 8-frame budget, GIFT improves the VideoMME overall score by 9.1\% (from 53.9 to 58.8) and surges the MLVU score by a remarkable 15.0\%. Crucially, GIFT's performance benefits remain significant even on more powerful baselines like Qwen2.5-VL and VideoLLaMA3. Specifically, GIFT elevates the LongVideoBench  score of the Qwen2.5-VL (8 frames) by 10.6\% (from 52.7 to 58.3), while the impact is even more striking on VideoLLaMA3 (8 frames), boosting its MLVU score from 59.1 to an impressive 70.7 (+19.6\%). These results underscore that GIFT can serve as a plug-and-play, model-agnostic module to enhance existing video models with minimal integration effort. Its core principle of identifying \emph{irreplaceability frame} provides a fundamentally more effective set of visual inputs, thereby unlocking the latent potential of a wide spectrum of VLMs.
\subsection{Ablation study}
\subsubsection{Ablation study on different modules}
\begin{figure}[t]
  \centering
   \includegraphics[width=1.0\linewidth]{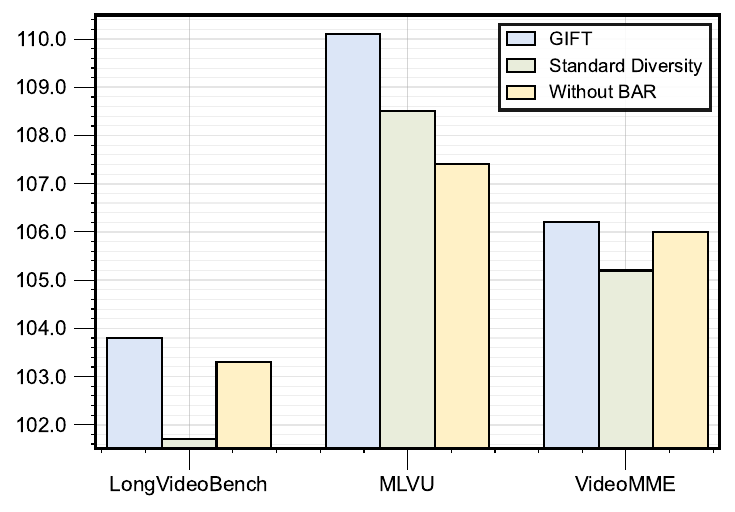}

   \caption{\textbf{Ablation study of GIFT's modules on LLaVA-Video (Frame Budget:32)}. The y-axis shows performance relative to Uniform Sampling (100\%). ``Standard Diversity'' refers to using the decoupled evaluation criteria with standard diversity instead of our proposed directed diversity. ``Without BAR'' refers to disabling the Budget-Aware Refinement module. }
   \label{fig:ablation}
   \vspace{-3mm}
\end{figure}

To dissect the contribution of our two key components, we perform ablation experiments on the LLaVA-Video model with a frame budget of 32. The results are presented in \cref{fig:ablation}, where the y-axis represents the performance score relative to the Uniform Sampling baseline (100\%).

\noindent \textbf{Effectiveness of Directed Diversity:} 
Replacing our Directed Diversity with a standard, undirected diversity metric (which calculates the mean distance to all other frames) leads to a significant performance degradation across all benchmarks. This is particularly evident on long-form video benchmarks like LongVideoBench, where performance drops from 103.8\% to 101.7\%. This reveals the inherent limitation of traditional, relevance-agnostic diversity metrics. By aiming to maximize visual separation from all other frames, they are designed to find what is merely visually novel, not what is truly informationally unique. This crucial distinction often causes them to select irrelevant noise frames. In contrast, our Directed Diversity is explicitly conditioned on query relevance. It is designed to preserve those frames that are unique relative to their potential substitutes, which is the key to its robust performance.

\noindent \textbf{Importance of Budget-Aware Refinement:} Disabling the Budget-Aware Refinement (BAR) strategy  results in a clear performance drop on LongVideoBench and MLVU by 0.5\% and 2.7\%, respectively. This failure underscores the inherent limitation of static mechanism. Without BAR, the strong suppressive effect from the initial selections remains fixed. This forces the algorithm to select isolated, low-relevance noise frames over more meaningful, yet suppressed, secondary frames that are crucial for temporal context. Our BAR component rectifies this by dynamically re-evaluating directed diversity at each stage, ensuring that additional budget is spent on building a richer coherent understanding of key events rather than on redundant or irrelevant content.
\begin{table}[t]
\caption{\textbf{Ablation study on the refinement batch size $B$ for our Budget-Aware Refinement (BAR) module}. Frames are selected iteratively in batches of $B$ in BAR module. Vanilla represents the performance of uniform sampling for comparison.}
\label{tab:3}
\centering
\renewcommand{\arraystretch}{1.05} 
\setlength{\tabcolsep}{1mm}
\begin{tabular}{c|cccccc}
\hline
\multirow{2}{*}{\makecell{$B$ in \\ BAR}}& \multirow{2}{*}{\textbf{LVB}}& \multicolumn{4}{c}{\textbf{VideoMME(w.o. sub.)}} & \multirow{2}{*}{average }\\
\cline{3-6}
 & & Overall & Short & Medium & Long & \\
\hline
Vanilla & 58.0& 62.6& 76.3& 59.2& 52.3& 60.3\\ 
\hline
$B=6$& 59.8& \textbf{65.2}& \textbf{78.1}& 63.0& \textbf{54.6}& \underline{62.5}\\
$B=7$& \underline{60.0}& 64.8& 76.7& 63.9& 53.9& 62.4
\\
$B=8$& 59.7& \underline{65.1}& 76.8& \textbf{64.8}& 53.9& 62.4
\\
$B=9$& \textbf{60.2}& 65.0& \underline{77.3}& 64.1& 53.6
& \textbf{62.6}\\
$B=10$& 59.0& \underline{65.1}& 76.8& \underline{64.6}& 53.9& 62.1\\
$B=11$& 59.3& 65.0& 77.0& 64.0& \underline{54.0}& 62.2\\
$B=12$& 58.9& 64.9& 77.1& \underline{64.6}& 53.0& 61.9\\
\hline
\end{tabular}
\vspace{-1mm}
\end{table}
\subsubsection{Ablation study on BAR}
To further investigate the impact of our Budget-Aware Refinement module, we conduct a sensitivity analysis on its core hyperparameter $B$. This parameter controls the granularity of our iterative  process: a smaller $B$ leads to more frequent re-evaluation stages. The results, presented in ~\cref{tab:3}, clearly demonstrate the effectiveness of our dynamic strategy, as all settings of $B$ significantly outperform the uniform sampling baseline. This confirms the fundamental effectiveness of our dynamic selection strategy. Within these settings, we observe that performance peaks at $B=9$, suggesting an optimal trade-off in refinement granularity. A small $B$ might over-prioritize building temporal context around the very first keyframes selected, potentially at the expense of capturing other distinct, high-relevance events. Conversely, a large $B$ makes the process resemble a one-shot selection, failing to effectively release the suppression on contextually valuable frames in time. Based on this analysis, we set $B=9$ as default for all other experiments.

\section{Conclusion}
\label{sec:conclusion}
In this paper, we presented GIFT, a training-free keyframe selection framework that substantially enhances the efficiency and effectiveness of various VLMs. Unlike prior methods that relied on myopic, greedy-based selection, GIFT reformulates frame selection from a global optimization perspective by quantifying each frame's irreplaceability via our proposed directed diversity. Subsequently, the Budget-Aware Refinement strategy adaptively adjusts the selection priority as the budget increases, progressively building rich temporal context. Extensive experimental results across multiple VLMs and benchmarks demonstrate that GIFT consistently improves model performance in different frame budgets, providing a practical solution for long-form video understanding.
\clearpage
{
    \small
    \bibliographystyle{ieeenat_fullname}
    \bibliography{main}
}


\end{document}